\documentclass[twocolumn, switch]{article}
\usepackage{preprint}
\SetBgContents{}    % preprint.sty draws corner tick marks; suppress them
\usepackage{amsmath, amsthm, amssymb, amsfonts, mathtools}
\usepackage[round, authoryear]{natbib}
\usepackage[utf8]{inputenc}
\usepackage[T1]{fontenc}
\usepackage{xcolor}
\usepackage[colorlinks=true,
            linkcolor=purple,
            urlcolor=blue,
            citecolor=cyan,
            anchorcolor=black]{hyperref}
\usepackage{booktabs}
\usepackage{microtype}
\usepackage{float}
\usepackage{dblfloatfix}    % let figure*/table* float to bottom of pages too
\usepackage{enumitem}

\usepackage{titlesec}
\titlespacing\section{0pt}{12pt plus 3pt minus 3pt}{1pt plus 1pt minus 1pt}
\titlespacing\subsection{0pt}{10pt plus 3pt minus 3pt}{1pt plus 1pt minus 1pt}
\titlespacing\subsubsection{0pt}{8pt plus 3pt minus 3pt}{1pt plus 1pt minus 1pt}

\graphicspath{{figs/}}

\title{Beyond the Smile:\\
A Hybrid Convolutional VAE for Crypto Volatility Surfaces}

\usepackage{authblk}

\author[1\thanks{Corresponding author:
\href{mailto:sadanand@jasper-research.example}{\tt sadanand@jasper-research.example}}]{Sadanand Singh}
\author[1]{Allam Reddy}
\author[1]{Manan Chopra}
\affil[1]{Jasper Research, USA}

\begin{document}

\twocolumn[
  \begin{@twocolumnfalse}

\maketitle

\begin{abstract}
We present a convolutional variational autoencoder for
cryptocurrency implied-volatility surfaces, together with a
deployable predictor that combines it with a quadratic smile re-fit
through a deterministic per-tenor routing rule. Trained on $6{,}034$
fully-filled hourly Binance Options surfaces of BTC and ETH spanning
May--October $2023$ and parameterised on a common $6 \times 7$
tenor--delta grid,
the model attains a hidden-cell surface-completion RMSE in the
$0.94$--$1.56$ vol-point range across both markets and mask rates
$10$--$50\%$. The hybrid
predictor attains $0.83$ vol points at $50\%$ masking against $7.00$
for the smile re-fit alone, an eightfold reduction obtained at no
additional inference cost. Under structurally-correlated hole
patterns that emulate the withdrawal of an entire tenor of strikes,
the smile re-fit incurs $9.6$--$13.1$ vol points of error while the
learned model remains at $1.5$--$1.9$, isolating a regime in which
the generative model is the only viable predictor. Joint training on BTC and ETH improves the in-distribution model on
both markets by $9$--$27\%$ relative to the better-performing
single-symbol counterpart, indicating a substantially shared
vol-surface manifold across the two largest cryptocurrencies over
the observation window. The hybrid is calendar- and
butterfly-arbitrage-free at the listed strikes, a property that the
parametric smile re-fit alone fails at high mask rates. The
per-snapshot reconstruction error of the trained model flags the
late-October ETF-anticipation rally and the August $17$, $2023$
flash crash as elevated-error periods without supervision. All training and evaluation infrastructure is
released to support reproducible follow-on work.
\end{abstract}

\vspace{0.4em}
\noindent\textbf{Keywords:} implied volatility surface; variational
autoencoders; cryptocurrency options; surface completion; cross-asset
transfer; anomaly detection.
\vspace{0.6em}
  \end{@twocolumnfalse}
]

% =============================================================================
\section{Introduction}
\label{sec:intro}

The implied-volatility (IV) surface is the primary state variable of an
options book. Risk systems, market-making engines, and structured-product
desks all rely on a continuous, well-behaved surface estimated at every
update, despite the fact that the observed chain is irregular, partially
quoted, and at any moment contains stale or absent strikes. The
practitioner's standard response is a smooth parametric smile model fit
per maturity (SVI \citep{gatheral2014svi}, SABR \citep{hagan2002sabr},
or local-polynomial variants), together with cross-tenor interpolation
rules. This works well when the chain is densely populated near the
money, and breaks down precisely in the operational scenarios for which
the surface is most needed: when a feed goes silent, when wing liquidity
dries up, or when a single maturity is delisted from the venue.

Variational autoencoders \citep{kingma2014vae, rezende2014stochastic}
offer a complementary tool. A low-dimensional latent representation
of a population of historical surfaces yields a prior over plausible
shapes that can be used both to reconstruct a surface from a partial
observation and to flag surfaces lying outside the learned manifold.
Existing applications to equity-index volatility surfaces
\citep[e.g.][]{ackerer2020deepsmoothing, bergeronlund2024, reddy2019}
have demonstrated that the smooth, near-stationary equity smile
admits an effective low-dimensional encoding.

Cryptocurrency options constitute a comparatively young microstructure
and present a distinct combination of opportunities and constraints.
Markets trade continuously without a daily settlement break, hourly
data is publicly available from venues such as Binance and Deribit, and
the listed strike grid is typically denser in the wings than equity-index
chains. At the same time, single-event regime transitions (exchange
failures, regulatory announcements, and macro liquidity shocks) can
shift the surface by several volatility points within an hour. The
combination of high-frequency cadence and strong cross-currency
comovement between BTC and ETH implied volatilities provides a setting
in which the out-of-distribution behaviour of a learned surface model
can be studied along several axes simultaneously: across mask patterns,
across calendar regimes, and across underlying assets.

\subsection{Contributions}

This paper presents an empirical study of a masked-input VAE for the
cryptocurrency implied-volatility surface, with an evaluation protocol
informed by the operational requirements of market-making and
risk-management systems. The contributions are as follows.

\begin{enumerate}[leftmargin=*]
    \item \textbf{End-to-end pipeline and release.} A reproducible
    processing pipeline converts the public Binance Options
    end-of-hour (EOH) archive into a fixed $6 \times 7$ tenor--delta
    grid suitable for grid-shaped neural architectures, with quality
    flags and provenance preserved at each stage
    (Section~\ref{sec:data}); code and per-run artifacts are released
    with the manuscript.

    \item \textbf{Convolutional VAE with a deterministic hybrid
    routing rule.} A $2$D-convolutional masked-input VAE for the
    $6 \times 7$ tenor--delta grid, combined with the practitioner's
    standard quadratic smile re-fit through a per-tenor routing rule
    (defer to the smile re-fit when a tenor row retains at least
    three observed cells; invoke the ConvVAE otherwise), attains
    $0.83$ vol points of completion RMSE at $50\%$ random masking
    against $7.00$ for the smile baseline alone, an eightfold
    reduction obtained at no additional inference cost
    (Section~\ref{sec:completion}). The architecture choice is
    justified by an internal ablation against MLP and self-attention
    encoder--decoders trained on the same data
    (Section~\ref{sec:arch}).

    \item \textbf{Failure-mode separation between parametric and
    learned predictors.} Under structured holes that emulate
    plausible operational failures (an entire tenor of strikes
    withdrawn, or a full delta column unquoted), the smile re-fit
    incurs an order of magnitude greater error than the ConvVAE
    ($9.6$--$13.1$ vs.\ $1.5$--$1.9$ vol points), establishing a
    regime in which a generative prior is the only viable predictor
    rather than an incremental improvement
    (Section~\ref{sec:structured}).

    \item \textbf{Static no-arbitrage compliance.} The deployed
    predictor is calendar- and butterfly-arbitrage-free at the seven
    listed strikes per tenor on both markets, inheriting the gridded
    data's compliance profile cell for cell; calendar compliance is
    enforced by a free $L_2$ post-projection
    ($\le 0.001$ vol-point RMSE impact) and butterfly compliance
    holds empirically without enforcement. The parametric smile
    re-fit, by contrast, admits a butterfly arbitrage on $38.9\%$
    ($33.1\%$) of BTC (ETH) reconstructions at $50\%$ masking
    (Section~\ref{sec:arb}).

    \item \textbf{Cross-asset transfer.} A ConvVAE trained on BTC
    alone attains within $5$--$27\%$ of its in-distribution accuracy
    when evaluated on ETH under the target's own normalisation. Joint
    training on BTC and ETH yields a further $9$--$27\%$ reduction
    relative to the better-performing single-symbol counterpart on
    both markets, indicating a substantially shared vol-surface
    manifold across the two largest cryptocurrencies over the
    observation window (Section~\ref{sec:cross_market}).

    \item \textbf{Unsupervised anomaly signal.} The per-snapshot
    reconstruction error of the trained ConvVAE, evaluated without
    masking, flags known dislocations (the late-October
    ETF-anticipation rally and the August $17$, $2023$ flash
    crash) as elevated-error periods without supervision, and the
    latent representation exhibits an interpretable temporal
    trajectory with anomalies concentrated at the manifold periphery
    (Section~\ref{sec:anomaly}).
\end{enumerate}

% =============================================================================
\section{Related Work}
\label{sec:related}

\paragraph{Parametric smile models.}
Parametric smile models have a long lineage in derivative pricing.
The stochastic-volatility-inspired (SVI) parameterisation
\citep{gatheral2014svi} and its arbitrage-free refinements, the SABR
model \citep{hagan2002sabr}, and lower-order polynomial
parameterisations remain the standard choice on equity-index desks;
\citet{gatheral2006vol} provides a comprehensive practitioner
treatment. Arbitrage-free smoothing of the empirical surface, prior
to any pricing application, is itself a substantial sub-literature
\citep{fengler2007arbitragefree}. We adopt the
quadratic-in-log-moneyness variant as our parametric baseline
because it is the inverse of the gridding procedure we use to
construct training targets (Section~\ref{sec:grid}).

\paragraph{Arbitrage-free smoothing.}
A second strand within the parametric tradition enforces static
no-arbitrage on the empirical surface independently of the choice of
smile family. \citet{fengler2007arbitragefree} constructs $C^2$
smoothing splines on call prices that automatically satisfy butterfly
and calendar conditions in strike space; \citet{gatheral2014svi}
characterise the arbitrage-free subset of the SVI parameter space and
provide explicit closed-form conditions on its coefficients;
\citet{ackerer2020deepsmoothing} embed analogous penalties as soft
constraints in a neural smoother and demonstrate large reductions in
violation rates on equity-index surfaces. Section~\ref{sec:arb}
adopts the discrete analogue of these conditions at the seven listed
strikes per tenor and reports compliance empirically for the
deployed predictor rather than enforcing it through a constrained
parameterisation.

\paragraph{Statistical decomposition of vol surfaces.}
\citet{cont2002dynamics} introduce principal-component (PCA) and
functional-PCA decompositions of the IV surface, identifying a small
number of orthogonal shape factors (level, skew, term-structure
slope, and curvature) that account for the bulk of empirical
variation. A PCA-based completion baseline (fit the leading
components on training surfaces, then solve for the latent code that
best matches the observed cells) is the closest classical analogue
of a VAE for our completion task and provides a non-neural
comparator (Section~\ref{sec:baselines}).

\paragraph{Neural surface modelling.}
A first generation of deep-learning work on volatility models has
focused on \emph{pricing} and \emph{calibration} under prescribed
stochastic processes \citep{horvath2021deep, bayer2019deep}. Such
approaches substitute a fitted neural function for expensive Monte
Carlo simulation and are largely orthogonal to the question of how
the empirical surface should be represented. A second line of work
models the surface directly: \citet{ackerer2020deepsmoothing}
propose a constrained neural smoother on SPX surfaces with explicit
butterfly and calendar no-arbitrage penalties, and demonstrate that
a learned representation outperforms local-polynomial baselines.
The generative-modelling perspective is initiated by
\citet{reddy2019}, who train a variational autoencoder on
single-maturity smiles generated by a SABR model, and is extended
to multi-maturity SPX surfaces by \citet{bergeronlund2024}.
Generative-adversarial formulations have been pursued for related
problems by \citet{cuchiero2020gan} and \citet{wiese2020quantgans};
we do not implement a GAN baseline here.

\paragraph{Masked-input training and architectural priors.}
The masked-input training paradigm we adopt is standard in
self-supervised representation learning, originating with the
context-encoder formulation of \citet{pathak2016context} and most
recently scaled in the masked autoencoder of \citet{he2022mae}.
The convolutional and self-attention architectures we compare in
Section~\ref{sec:arch} correspond to two well-established families
of inductive bias for grid-structured inputs: locality and
translation equivariance for the former, and
permutation-equivariant pairwise interaction for the latter,
building on the original transformer \citep{vaswani2017attention}
and its set-structured variant \citep{lee2019settransformer}.

\paragraph{Anomaly detection.}
The use of reconstruction error from a generative model as an
\emph{anomaly score} is well-established
\citep{an2015vaeanomaly, ruff2021review}. The analysis in
Section~\ref{sec:anomaly} applies this approach to cryptocurrency
IV surfaces using the per-snapshot reconstruction error of an
unmasked input as an unsupervised statistic, and is presented as an
analytical by-product of the completion model rather than as a
methodological contribution to the anomaly-detection literature.

\paragraph{Cryptocurrency options.}
\citet{madan2019bitcoin} and \citet{hou2020pricing} calibrate
parametric stochastic-vol models to Bitcoin options;
\citet{alexander2023crypto} document the BTC smile's index-option
characteristics. To our knowledge no prior work applies a
masked-input VAE jointly to BTC and ETH or examines the
parametric--learned routing policy that we identify as the
deployable configuration.

% =============================================================================
\section{Data}
\label{sec:data}

\subsection{Source}

Our primary dataset is the public Binance Options end-of-hour summary
archive,\footnote{\url{https://data.binance.vision/data/option/daily/EOHSummary/}}
which publishes one CSV per (symbol, date) containing $24$ hourly
snapshots of the full option chain together with best bid/ask prices,
sizes, open-interest, venue-supplied implied volatilities, and Greeks.
Coverage is May $18$, $2023$ through October $23$, $2023$, a span of $147$ days,
after which the archive was discontinued by the publisher. We use the
two most liquid pairs, BTCUSDT and ETHUSDT. Raw row counts and snapshot
counts are summarised in Table~\ref{tab:data_raw}.

\begin{table}[t]
\centering
\caption{Raw Binance Options EOH dataset. ``Snapshots'' is the number
of hourly observation timestamps. ``Avg options'' is the mean number
of listed option contracts observed per snapshot.}
\label{tab:data_raw}
\small
\begin{tabular}{lrrrr}
\toprule
Symbol  & Rows  & Days  & Snaps. & Avg opts/snap \\
\midrule
BTCUSDT & 796{,}972 & 147 & 3{,}501 & 226 \\
ETHUSDT & 979{,}331 & 147 & 3{,}501 & 278 \\
\bottomrule
\end{tabular}
\end{table}

\subsection{Cleaning, forward inversion, and quality flags}

We parse the contract names to extract expiry, strike, and option
right; coerce numeric columns; compute days-to-expiry and time-to-expiry
in years; and drop rows with non-positive DTE, malformed strike,
missing or zero mark price, or invalid right. The drop rate at this
stage is approximately $2.2\%$.

For each (snapshot, expiry) group we pair every (strike, call mid, put
mid) triplet and recover an implied forward via put--call parity
\citep{put_call_parity},
\begin{equation}
F = K + (C_{\text{mid}} - P_{\text{mid}}), \quad
\hat F_{\text{snap, exp}} = \operatorname{median}_K F(K),
\end{equation}
taking the median across strikes for robustness and assuming the USDT
short-rate $r \approx 0$ over the relevant tenors. Forward coverage is
$99.5$--$99.7\%$ of rows. Implied vols on bid, ask, and mark prices are
recovered by Newton iteration on the Black--76 forward form
\citep{black1976}, warm-started from the venue-published
\texttt{mark\_iv}. The warm-start is the principal numerical
consideration: the published mark IV typically lies within $\pm 20\%$
of the bid/ask IV, which secures convergence of Newton's method in two
or three iterations.

After IV recovery we tag each row with quality flags \texttt{is\_quoted},
\texttt{is\_oi\_positive}, \texttt{is\_tight\_quote}
(half-spread $/ \text{mid} < 0.5$), \texttt{is\_train\_tenor}
($7 \le \text{DTE} \le 365$), \texttt{is\_train\_moneyness}
($K/F \in [0.5, 2.0]$), \texttt{is\_iv\_sane}
($\sigma \in [10\%, 300\%]$), and their boolean conjunction
\texttt{is\_train\_grade}. No rows are dropped from the cleaned
parquet; filters become columns so that the same cleaned data can be
re-used under different downstream criteria. Train-grade rows are
$52.2\%$ of BTC and $48.4\%$ of ETH.

We retain the venue-published \texttt{mark\_iv} as our canonical IV.
Our own inverted IVs from parity-derived forwards agree on the median
strike to $2.7$--$2.9$ vol points and disagree on the deep OTM tail by up
to $22$ vol points, which we attribute to the venue using its
perpetual-index price as the forward (rather than parity). We retain
our inverted bid/ask IVs as auxiliary features and treat
\texttt{mark\_iv} as the canonical value for downstream gridding.

\subsection{Smile re-fit and the $6\times 7$ grid}
\label{sec:grid}

For each snapshot and each listed expiry $T_i$ with at least five
train-grade strikes, we fit a total-variance quadratic in log-moneyness
$k = \log(K / F)$,
\begin{equation}
w_{T_i}(k) = a_i + b_i k + c_i k^2,
\qquad w(k) = \sigma^2(k)\, T,
\label{eq:quadratic}
\end{equation}
by ordinary least squares on the observed $(k, w)$ pairs. This
three-parameter form captures level, skew, and curvature, and is
deliberately less expressive than SVI; we adopt it both as the
parametric baseline and as the generator of the
$(\tau, \delta)$-grid. For each target tenor $T^{\star}$ we
linearly interpolate $(a, b, c)$ between the two bracketing fitted
expiries (equivalent to a linear-in-$T$ interpolation of total
variance at fixed $k$), with flat extrapolation at the boundaries.
For each target call delta $\delta$ we then solve, by fixed-point
iteration,
\begin{equation}
d_1 \coloneqq \frac{-k + \tfrac{1}{2} \sigma^2 T^{\star}}{\sigma \sqrt{T^{\star}}}
         = \Phi^{-1}(\delta),
\quad \sigma = \sqrt{w(k) / T^{\star}},
\label{eq:k_delta}
\end{equation}
iterating
$k \leftarrow \tfrac{1}{2}\sigma^2 T^{\star} - \Phi^{-1}(\delta)\,\sigma\sqrt{T^{\star}}$
to convergence in at most $50$ iterations. We reject any cell whose
converged $(k,\sigma)$ does not reproduce $\delta$ to within $10^{-4}$.

The target tenors and deltas are
\begin{equation*}
\tau \in \{14, 30, 60, 90, 120, 180\}\,\text{days},
\end{equation*}
\begin{equation*}
\delta \in \{0.10, 0.20, 0.30, 0.50, 0.70, 0.80, 0.90\},
\end{equation*}
giving $6 \times 7 = 42$ cells per surface. These specific values are
not canonical: they are the largest subset of the standard
$(7, 14, 30, 60, 90, 180, 365) \times (0.05, 0.10, 0.20, 0.30, 0.50,
0.70, 0.80, 0.90, 0.95)$ grid for which Binance's listings admit a
high fill rate. The archive contains effectively no
$365$-day listings in our window; $7$-day tenors are only
intermittently bracketed by listed expiries; and the $\delta = 0.05$
and $\delta = 0.95$ wings require long-range smile extrapolation that
is unreliable at long tenors. Restricting the grid to the corners
supported by the listed strikes raises the fully-filled-snapshot rate
from $9\%$ on the canonical grid to $80.9\%$ for BTC and $92.1\%$ for
ETH (Table~\ref{tab:data_grid}).

\begin{table}[t]
\centering
\caption{Gridded surfaces: counts of fully-filled hourly snapshots (no
NaN across all 42 cells) and per-cell IV coverage in the long-format
output.}
\label{tab:data_grid}
\small
\begin{tabular}{lrrrr}
\toprule
Symbol  & Snaps. & Fully filled & Pct. & Per-cell cov. \\
\midrule
BTCUSDT & 3{,}489 & 2{,}821 & $80.9\%$ & $99.8\%$ \\
ETHUSDT & 3{,}487 & 3{,}213 & $92.1\%$ & $99.8\%$ \\
\bottomrule
\end{tabular}
\end{table}

\subsection{Splits and per-symbol normalisation}

We adopt a single \emph{time-ordered} $70/15/15$ split of the
fully-filled snapshots, with train, validation, and test windows
formed from contiguous blocks of calendar time. This yields
$1{,}974 / 423 / 424$ snapshots for BTC and $2{,}249 / 481 / 483$ for
ETH. A random i.i.d.\ split would substantially overstate
out-of-sample accuracy in the presence of strong hour-to-hour
autocorrelation; the time-ordered split is the appropriate evaluation
unit and underlies every reported figure.

For each symbol we compute per-cell mean and standard deviation on the
corresponding training block and use these statistics to $z$-normalise
all inputs at both training and inference, including in the joint
training setting where two independently-normalised symbols are
concatenated. This procedure separates IV level from IV shape and
ensures that the cross-symbol experiments measure transfer of
\emph{shape} rather than coincidence of absolute levels.

% =============================================================================
\section{Methodology}
\label{sec:method}

\subsection{Masked-input ConvVAE}
\label{sec:convvae}

Let $X \in \mathbb{R}^{6 \times 7}$ be a $z$-normalised surface
arranged on the (tenor, delta) grid, and $M \in \{0, 1\}^{6 \times 7}$
the corresponding observation mask, with $M_{r,c} = 1$ indicating
that cell $(r, c)$ is observed and $M_{r,c} = 0$ that it is hidden.
We present the masked input to the encoder as a $2$-channel image
$[X \odot M;\, M] \in \mathbb{R}^{2 \times 6 \times 7}$. The encoder
is a stack of three $3 \times 3$ convolutional layers with GELU
activations, padded to preserve the $6 \times 7$ spatial dimensions
throughout; the resulting feature map is flattened and projected by
two linear heads to $(\boldsymbol\mu, \log \boldsymbol\sigma^2)$ on
$\mathbb{R}^z$. We use the standard reparameterisation
\citep{kingma2014vae},
$z = \boldsymbol\mu + \boldsymbol\sigma \odot \boldsymbol\varepsilon$
with $\boldsymbol\varepsilon \sim \mathcal{N}(\mathbf{0}, I)$ during
training and $z = \boldsymbol\mu$ at inference. The decoder is the
encoder's mirror image: a linear map from $\mathbb{R}^z$ to an
$h \times 6 \times 7$ feature map, three $3 \times 3$ convolutional
layers with GELU activations, and a final $1 \times 1$ projection
to the scalar output channel.

The convolutional architecture imposes two inductive biases
well-matched to the volatility surface: translation equivariance,
so that the same response is produced by a given local shape
feature regardless of where it occurs on the grid; and locality,
so that each layer's receptive field is restricted to a
$3 \times 3$ neighbourhood of cells before composition. Both are
consistent with the smooth tenor- and delta-wise structure of
empirical IV surfaces. Section~\ref{sec:arch} provides a direct
empirical comparison against alternative encoder--decoder families
that justifies this design.

The training objective is the masked $\beta$-VAE loss
\citep{kingma2014vae, higgins2017betavae}, weighted to favour
hidden-cell reconstruction:
\begin{equation}
\begin{aligned}
\mathcal{L}(x, m) &= w_{\mathrm{hid}}\,\bar{\ell}_{\bar m}(x, \hat x)
                   + w_{\mathrm{obs}}\,\bar{\ell}_{m}(x, \hat x) \\
                  &\quad + \beta \,\operatorname{KL}\big(q(z \mid x, m)\,\|\,\mathcal{N}(\mathbf{0}, I)\big),
\end{aligned}
\label{eq:loss}
\end{equation}
where $\bar m = 1 - m$ is the hidden-cell mask,
$\bar{\ell}_{\mathbf{w}}(x, \hat x) = \tfrac{1}{\|\mathbf{w}\|_1}
\sum_i w_i (x_i - \hat x_i)^2$ is the per-sample mean squared error
over the cells flagged by $\mathbf{w}$, and the KL term is computed in
closed form. Unless noted otherwise we set $w_{\mathrm{hid}} = 1.0$,
$w_{\mathrm{obs}} = 0.1$, and $\beta = 10^{-3}$. The dominant weight
on hidden cells reflects that masked completion is the primary
objective; a small weight on observed cells maintains consistency
between the encoder's effective input and the decoder's reconstruction;
and a small $\beta$ avoids posterior collapse on a dataset whose
information content does not warrant strong regularisation of the
latent prior.

\subsection{Masking schemes}

At training time we draw, for each minibatch sample, a per-cell mask
in which a fixed number $n_{\mathrm{hid}} = \lfloor r \cdot 42 \rfloor$
of the $42$ cells are hidden, with $r \sim \mathcal{U}(0.10, 0.50)$
drawn independently per sample. We refer to this as the \emph{random}
mask scheme.

For the structured-hole experiments in Section~\ref{sec:structured} we
also consider five fixed-pattern masks:
\begin{itemize}[leftmargin=*]
    \item \texttt{row\_random}: one tenor row chosen uniformly at
    random is fully hidden ($7$ cells, $16.7\%$);
    \item \texttt{col\_random}: one delta column chosen uniformly at
    random is fully hidden ($6$ cells, $14.3\%$);
    \item \texttt{wing\_put}: deltas $\{0.10, 0.20\}$ for all tenors
    are hidden ($12$ cells, $28.6\%$);
    \item \texttt{wing\_call}: deltas $\{0.80, 0.90\}$ for all tenors
    are hidden ($12$ cells, $28.6\%$);
    \item \texttt{long\_tenor}: the $180$d row is hidden ($7$ cells,
    $16.7\%$).
\end{itemize}

\subsection{Smile re-fit baseline}

The baseline inverts the gridding procedure described in
Section~\ref{sec:grid}, applied to the observed cells of a held-out
surface. For each tenor row containing at least three observed cells
we re-fit the quadratic in Eq.~\eqref{eq:quadratic} on the observed
$(k_{\text{cell}}, \sigma_{\text{cell}})$ pairs by least squares.
Tenors with fewer than three observations inherit smile parameters by
linear interpolation across neighbouring fitted tenors. For each
hidden cell we then solve Eq.~\eqref{eq:k_delta} at that cell's
target delta and report the implied $\sigma$.

The baseline is deliberately strong: the gridded targets defined in
Section~\ref{sec:grid} are themselves the output of exactly this
procedure applied to the full observed chain. When each tenor row is
sufficiently populated, the re-fit asymptotes to the inverse of the
data-generating map and the RMSE on hidden cells approaches zero. We
regard this not as a coincidence but as the appropriate parametric
oracle: any learned model must demonstrate value in the regime where
the parametric family is insufficient, not where it is.

\subsection{Hybrid routing rule}
\label{sec:hybrid}

We define a deterministic routing rule that combines the two
predictors:
\begin{equation}
\hat \sigma^{\mathrm{hybrid}}_{r,c}(x, m) =
\begin{cases}
\hat \sigma^{\mathrm{refit}}_{r,c}(x, m) & \text{if}\; \sum_{c'} m_{r,c'} \ge 3, \\
\hat \sigma^{\mathrm{ConvVAE}}_{r,c}(x, m) & \text{otherwise.}
\end{cases}
\label{eq:hybrid}
\end{equation}
The threshold of three follows from the rank requirement of the
quadratic in Eq.~\eqref{eq:quadratic}: three is the minimum number of
distinct $(k, w)$ pairs required to identify $(a, b, c)$. Below this
threshold no per-tenor re-fit is well-posed; above it the re-fit
attains near-optimal accuracy on the gridded targets. The rule is
therefore not a learned ensemble but a deterministic decomposition
grounded in the rank deficiency of the parametric model.

\subsection{Architecture and training details}

All experiments follow the architecture and optimiser settings of
Table~\ref{tab:hparams}, varying only the dimensions explicitly
ablated. Implementation is in PyTorch \citep{pytorch2019}; a single
configuration trains in $10$--$50$ seconds on a consumer-grade GPU,
and the complete ablation grid reported below trains and evaluates
within five minutes of wall-clock time. Per-run configurations,
training histories, checkpoints, and evaluation metrics are stored in
immutable per-run directories and released with the manuscript.

\begin{table}[!htbp]
\centering
\caption{Default hyperparameters. Where a sweep is reported we vary
the indicated entry and hold the rest fixed.}
\label{tab:hparams}
\small
\begin{tabular}{ll}
\toprule
Hyperparameter & Default \\
\midrule
Architecture & ConvVAE ($3 \times 3$ conv, 3 layers) \\
Latent dim $z$ & $16$ \\
Channel width $h$ & $64$ \\
Activation & GELU (encoder + decoder) \\
$w_{\mathrm{hid}}, w_{\mathrm{obs}}, \beta$ & $1.0,\ 0.1,\ 10^{-3}$ \\
Optimiser & Adam \citep{kingma2015adam}, lr $=10^{-3}$ \\
Batch size & $128$ \\
Epochs & $500$ (max), patience $50$ \\
Mask rate (random) & $r \sim \mathcal{U}(0.10, 0.50)$ per sample \\
Splits & $70/15/15$, time-ordered \\
\bottomrule
\end{tabular}
\end{table}

% =============================================================================
\section{Experimental Setup}
\label{sec:setup}

We report results in \emph{vol points} (i.e.\ $0.01 = 1$ vol pt) so
that errors can be read against the absolute IV level (BTC ATM IV is
in the $25$--$60\%$ range in our window). All RMSEs are computed on
the \emph{hidden} cells of the masked input only: observed cells are
returned unchanged by both the baseline and the hybrid, so including
them would artificially deflate error rates and produce a misleading
comparison.

For each ablation we evaluate at five mask rates
$r \in \{0.10, 0.20, 0.30, 0.40, 0.50\}$ on the same fixed-seed masks
across methods, so direct comparison between methods at the same mask
rate is paired. Where appropriate we additionally report results on
the five structured-hole schemes of Section~\ref{sec:method}. The
smile-re-fit baseline numbers are computed once per test set and
cached, since they depend only on the data and the mask seed, not on
any learned parameter.

% =============================================================================
\section{Architecture Selection}
\label{sec:arch}

The convolutional encoder--decoder of Section~\ref{sec:convvae} is one
of several natural choices for a $42$-cell grid. We motivate it
empirically by comparing it against two alternatives that bracket it
in terms of inductive bias.

\paragraph{MLP baseline.}
A fully-connected MLP that ignores the $(\text{tenor}, \text{delta})$
layout entirely: a two-layer encoder MLP with hidden width $128$ and
GELU activations operating on the flattened concatenation
$[x \odot m;\, m] \in \mathbb{R}^{84}$, with linear heads producing
$(\mu, \log\sigma^2)$, and a symmetric two-layer decoder. The MLP has
no spatial prior whatsoever.

\paragraph{AttnVAE.}
A permutation-equivariant encoder representing the surface as a set of
$42$ tokens. Each cell token receives a learned
$(\text{tenor},\,\text{delta})$ positional embedding together with its
(masked value, mask flag). Two multi-head self-attention layers with
feed-forward sub-layers update the tokens; the encoder pools the cell
representations by mean before projecting to $(\mu, \log\sigma^2)$.
The decoder broadcasts the latent across cell positions, applies a
symmetric attention stack, and projects each token to a scalar value.

\begin{figure*}[!htbp]
\centering
\includegraphics[width=\linewidth]{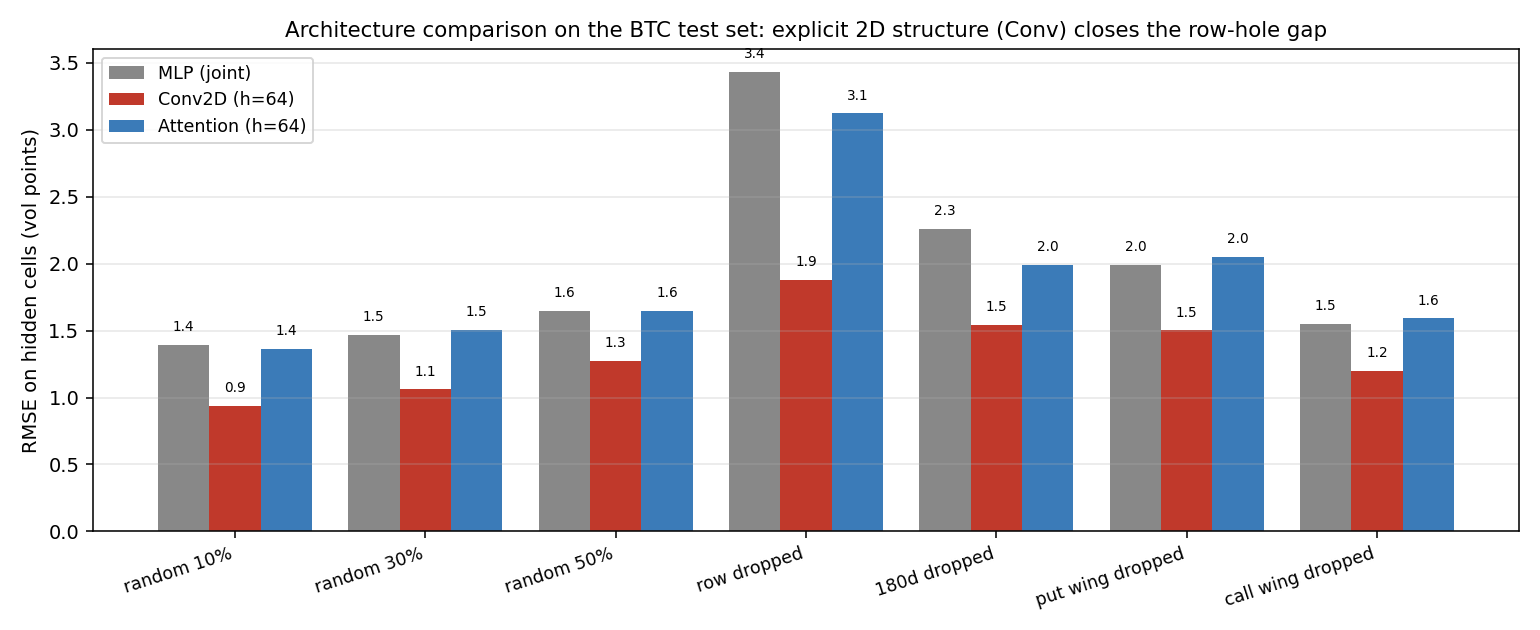}
\caption{Architecture comparison on the BTC test set across the seven
evaluation scenarios. The ConvVAE attains the lowest error in every
scenario; the largest absolute reduction relative to the MLP is on
the row-shaped structured holes (\texttt{row\_random},
$3.43 \to 1.88$, a $45\%$ reduction). The AttnVAE matches the MLP at
random masking and is uniformly worse than the ConvVAE despite a
comparable parameter budget, indicating that the locality bias of
the convolutional kernels, not raw capacity, is what carries the
result.}
\label{fig:arch}
\end{figure*}

\paragraph{Comparison protocol.}
All three architectures are trained on the joint BTC$+$ETH
$z$-normalised training set ($4{,}223$ surfaces) under identical loss
weights, optimiser, mask schedule, and epoch budget; they differ only
in the encoder--decoder. We hold the latent dimension fixed at
$z = 16$. The MLP uses hidden width $128$ ($56$k parameters); the
ConvVAE uses $64$ feature channels ($318$k parameters); the AttnVAE
uses $64$ token dimensions across two layers ($221$k parameters).
Table~\ref{tab:arch} reports BTC-test RMSE on three random-mask
rates and the four structured-hole patterns of
Section~\ref{sec:method}; Figure~\ref{fig:arch} visualises the same
comparison.

\begin{table}[!htbp]
\centering
\caption{Architecture comparison on the BTC test set: hidden-cell
RMSE (vol points) for an MLP encoder--decoder, a $2$D-convolutional
encoder--decoder (ConvVAE), and a per-cell self-attention
encoder--decoder (AttnVAE). All models are trained on the joint
BTC$+$ETH set with $z = 16$ and the same loss, optimiser, and mask
schedule; they differ only in the encoder--decoder. Lowest entry per
row in bold.}
\label{tab:arch}
\small
\setlength{\tabcolsep}{4.5pt}
\begin{tabular}{lccc}
\toprule
Scenario & MLP & ConvVAE & AttnVAE \\
\midrule
random $10\%$       & 1.40 & \textbf{0.94} & 1.36 \\
random $30\%$       & 1.51 & \textbf{1.07} & 1.55 \\
random $50\%$       & 1.67 & \textbf{1.25} & 1.68 \\
\midrule
\texttt{row\_random}   & 3.43 & \textbf{1.88} & 3.12 \\
\texttt{long\_tenor}   & 2.26 & \textbf{1.54} & 1.99 \\
\texttt{wing\_put}     & 1.99 & \textbf{1.50} & 2.05 \\
\texttt{wing\_call}    & 1.55 & \textbf{1.20} & 1.59 \\
\midrule
Parameters     & $56$k & $318$k & $221$k \\
\bottomrule
\end{tabular}
\end{table}

The ConvVAE attains the lowest error in every evaluated scenario,
with random-mask reductions of $25$--$33\%$ relative to the MLP and
a $45\%$ reduction on the \texttt{row\_random} structured hole. The
locality bias of the $3 \times 3$ kernels accounts for the
row-shaped-hole result: when one row is unobserved the
convolutional receptive field still pools information from the two
adjacent rows in the first layer, and from progressively wider
neighbourhoods thereafter, whereas an MLP must allocate capacity to
discover the same relationship from a flat representation. The
AttnVAE underperforms the ConvVAE at every scenario and matches
only the MLP at random masking, despite a comparable parameter
budget ($221$k vs.\ $318$k). The discrepancy between the two
parameter-heavy models points to inductive bias rather than raw
capacity: the convolutional locality and translation-equivariance
priors capture the dominant tenor- and delta-wise smoothness much
more efficiently than full self-attention does at our $4{,}223$-surface
training scale. We report ConvVAE results in the remainder of the
paper.

% =============================================================================
\section{Surface Completion}
\label{sec:completion}

\begin{table}[!htbp]
\centering
\caption{Hybrid completion RMSE (vol points) on the BTC test set vs.\
its components. The ConvVAE is the joint-trained $z{=}16$, $h{=}64$
model. Lowest entry per column in bold.}
\label{tab:hybrid}
\small
\begin{tabular}{lccccc}
\toprule
Method & $0.10$ & $0.20$ & $0.30$ & $0.40$ & $0.50$ \\
\midrule
Smile re-fit only       & \textbf{0.00} & 1.28 & 2.53 & 4.05 & 7.00 \\
ConvVAE only (joint)    & 0.94 & 1.01 & 1.07 & 1.13 & 1.25 \\
\textbf{Hybrid}         & \textbf{0.00} & \textbf{0.12} & \textbf{0.32} & \textbf{0.50} & \textbf{0.83} \\
\bottomrule
\end{tabular}
\end{table}

\begin{figure}[!htbp]
\centering
\includegraphics[width=\linewidth]{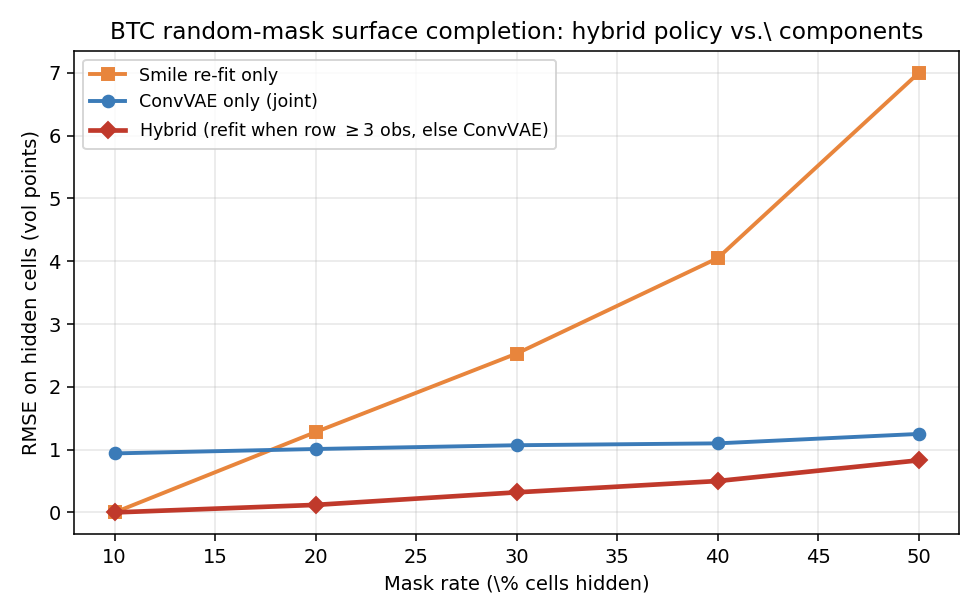}
\caption{Surface completion: the hybrid routing rule (red diamonds)
attains lower hidden-cell RMSE than either component at every random
mask rate on the BTC test set. At $50\%$ masking the hybrid ($0.83$
vol points) is more than $8\times$ more accurate than the smile
re-fit alone and a third more accurate than the ConvVAE alone.}
\label{fig:hybrid}
\end{figure}

The two component predictors exhibit complementary error profiles.
The smile re-fit is rank-sufficient and near-optimal at low mask
rates, where most tenor rows retain enough observed cells to identify
the quadratic in Eq.~\eqref{eq:quadratic}, but its error grows
roughly linearly with mask rate as a growing fraction of rows lose
rank. The ConvVAE attains a nearly mask-rate-independent accuracy in
the range $0.94$--$1.25$ vol points by drawing on the learned
manifold rather than re-solving a parametric fit. The routing rule
of Eq.~\eqref{eq:hybrid} composes them so that at every random mask
rate the hybrid is at least as accurate as the better component:
it attains the smile re-fit's near-zero error at low masking and
inherits the ConvVAE's robust accuracy at high masking
(Table~\ref{tab:hybrid}, Figure~\ref{fig:hybrid}). At $50\%$ random
masking the hybrid attains $0.83$ vol points, a more-than-eightfold
reduction relative to the smile re-fit alone and a $34\%$ reduction
relative to the ConvVAE alone. The rule introduces no additional
model parameters and no latency beyond what the two components
already require.

\begin{figure*}[t]
\centering
\includegraphics[width=\linewidth]{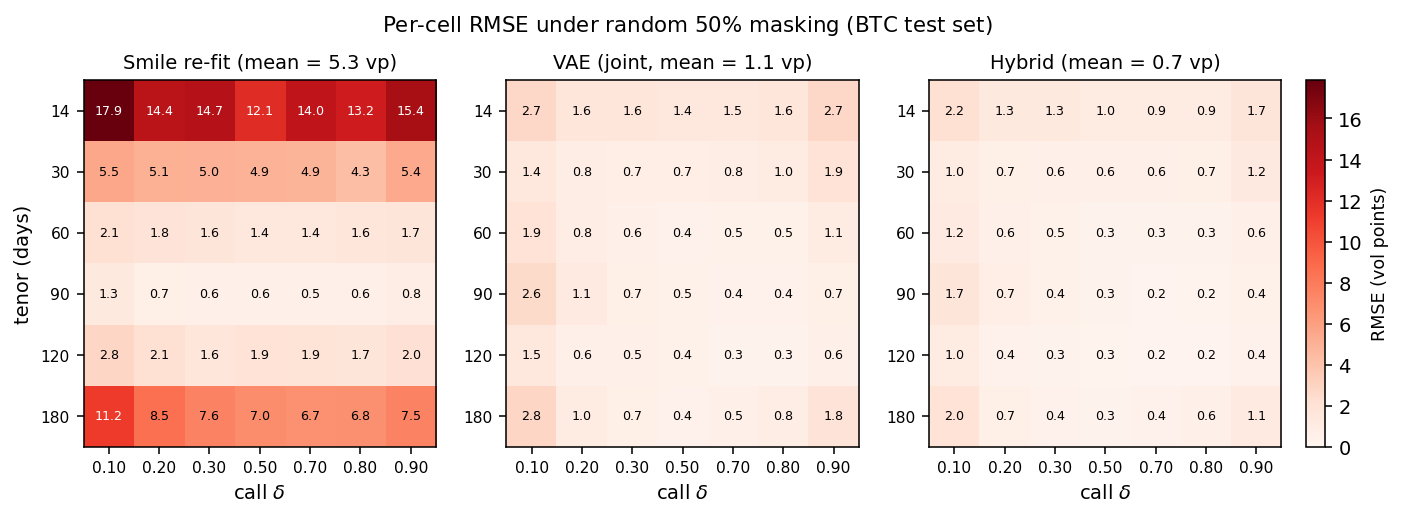}
\caption{Per-cell RMSE on the BTC test set under random $50\%$
masking, evaluated only on cells that were hidden. The smile re-fit
(mean $5.3$ vol points) incurs its largest errors at the boundary
tenors ($14$d and $180$d); the joint ConvVAE (mean $1.1$ vol
points) distributes its smaller errors more uniformly; the hybrid
(mean $0.7$ vol points) attains the lowest per-cell error in
nearly every cell by routing each hidden cell to the predictor that
is structurally qualified for it.}
\label{fig:residual_maps}
\end{figure*}

Figure~\ref{fig:residual_maps} resolves the aggregate RMSE to the
individual cells of the $6 \times 7$ grid at $r = 0.5$. The smile
re-fit incurs its largest per-cell errors at the boundary tenors
($14$d and $180$d), where cross-tenor extrapolation is supported by
a single neighbour rather than two and the deep-wing cells at the
$14$d row exceed $17$ vol points. The ConvVAE distributes
substantially smaller errors more uniformly across the grid. The
hybrid is below either component in nearly every cell, because the
per-tenor routing rule selects whichever predictor is structurally
qualified for each hidden cell, preserving the re-fit's near-zero
error at well-populated tenors and substituting the ConvVAE only
where rank deficiency forces it.

\subsection{Comparison against published baselines}
\label{sec:baselines}

The hybrid result above is established against the practitioner's
parametric oracle. Table~\ref{tab:baselines} places it within a
broader set of baselines: a PCA decomposition in the style of
\citet{cont2002dynamics}, an Ackerer-style deterministic deep
smoother with a soft calendar-arbitrage penalty
\citep{ackerer2020deepsmoothing}, and the joint-trained ConvVAE.
All learned models are trained on identical data with identical
splits and mask schedule; the smile re-fit and PCA require no
training.

\begin{table*}[!htbp]
\centering
\caption{Surface-completion RMSE on the BTC test set (vol points)
across baselines and the proposed configuration. PCA is reported at
its best operating point ($k = 8$ principal components, selected by
random-mask average); ``Deep Smoothing'' is an amortised
Ackerer-style deterministic autoencoder with a calendar-arbitrage
penalty; ConvVAE and Hybrid are the configurations of this paper.
Lowest entry per column in bold.}
\label{tab:baselines}
\vspace{6pt}
\small
\setlength{\tabcolsep}{8pt}
\begin{tabular}{lccccc}
\toprule
 & \multicolumn{3}{c}{Random mask} & \multicolumn{2}{c}{Row-shaped hole} \\
\cmidrule(lr){2-4} \cmidrule(lr){5-6}
Method & $r = 0.10$ & $r = 0.30$ & $r = 0.50$ & \texttt{row\_random} & \texttt{long\_tenor} \\
\midrule
Smile re-fit  & \textbf{0.00} & 2.53 & 7.00 & 9.59 & 13.12 \\
PCA, $k=8$ \citep{cont2002dynamics}    & 1.06 & 1.21 & 1.40 & 4.15 & 3.51 \\
Deep Smoothing \citep{ackerer2020deepsmoothing} & 0.99 & 1.16 & 1.38 & 2.30 & 2.26 \\
ConvVAE (ours) & 0.94 & 1.07 & 1.25 & \textbf{1.88} & \textbf{1.54} \\
\midrule
\textbf{Hybrid (ours)} & \textbf{0.00} & \textbf{0.32} & \textbf{0.83} & \textbf{1.88} & \textbf{1.54} \\
\bottomrule
\end{tabular}
\end{table*}

Three observations follow from Table~\ref{tab:baselines}. The
ConvVAE is the lowest-error \emph{learned} predictor on every column.
Against the PCA decomposition the gain is modest at random masks
($10$--$12\%$ reduction across the three rates evaluated) but
substantial on row-shaped holes ($55\%$ on \texttt{row\_random}),
reflecting the inability of a linear subspace to recover the
across-tenor dependence required when a full row is unobserved.
Against the Ackerer-style deep smoother the gain is $5$--$9\%$ at
random masks and $18$--$32\%$ on row-shaped holes, attributable to
the explicit two-dimensional grid structure of the ConvVAE that the
smoother's flat MLP lacks. No learned model beats the smile re-fit
at low mask rates, since the gridded targets are parametric in that
regime; the hybrid routing rule exploits this. Finally, none of the
baselines (parametric, statistical, or neural) approaches the
hybrid at the operating points where the hybrid dominates: at $50\%$
random masking the hybrid attains $0.83$ vol points against the
next-best $1.25$ for the ConvVAE alone and $1.38$ for the deep
smoother.

\subsection{Static no-arbitrage compliance}
\label{sec:arb}

A frequent objection to neural surface models is the absence of
static no-arbitrage guarantees. We test this empirically against
both standard conditions on the seven listed strikes per tenor.

\paragraph{Calendar arbitrage.}
The calendar condition in the delta parameterisation requires that
total variance $w(T_i, \delta_j) = \sigma^2(T_i, \delta_j)\, T_i$ be
non-decreasing in $T_i$ at fixed $\delta_j$
\citep{ackerer2020deepsmoothing}. We project per delta column by
$L_2$ isotonic regression (Pool-Adjacent-Violators) and compare
hidden-cell RMSE before and after.

\paragraph{Butterfly arbitrage.}
For each tenor row we recover the strike $K_{i,j} = F_i \exp(k_{i,j})$
of each cell from its vol via $k_{i,j} = \sigma_{i,j}^2 T_i / 2
- \sigma_{i,j} \sqrt{T_i}\, \Phi^{-1}(\delta_j)$, sort by $K$, and
check that the Black-76 forward call price $C(K_{i,j})/F_i =
\Phi(d_1) - \exp(k) \Phi(d_2)$ is convex in $K$ by second divided
differences. Convexity is the discrete butterfly-arbitrage condition
at the listed strikes and requires no continuous-smile fit.

\begin{table}[!htbp]
\centering
\caption{Static no-arbitrage compliance at $r = 0.5$ random masking.
``Cal.''\ is the fraction of test surfaces with any calendar
violation in total variance; ``Bfly.''\ is the fraction with any
butterfly violation at a listed strike; ``$\Delta$RMSE''\ is the
change in hidden-cell RMSE (vol points) after calendar projection.}
\label{tab:arb}
\small
\setlength{\tabcolsep}{6pt}
\begin{tabular}{llccc}
\toprule
Market & Method & Cal. & Bfly. & $\Delta$RMSE \\
\midrule
BTC & ConvVAE       & $0.0\%$ & $0.0\%$  & $+0.000$ \\
BTC & Hybrid        & $0.0\%$ & $0.0\%$  & $+0.000$ \\
BTC & Smile re-fit  & $9.7\%$ & $38.9\%$ & $-0.125$ \\
BTC & \emph{data}   & $0.0\%$ & $0.0\%$  & n/a \\
\midrule
ETH & ConvVAE       & $0.2\%$ & $0.0\%$  & $+0.000$ \\
ETH & Hybrid        & $0.4\%$ & $0.0\%$  & $+0.000$ \\
ETH & Smile re-fit  & $9.1\%$ & $33.1\%$ & $-0.188$ \\
ETH & \emph{data}   & $0.2\%$ & $0.0\%$  & n/a \\
\bottomrule
\end{tabular}
\end{table}

Table~\ref{tab:arb} supports three conclusions. First, the
ConvVAE matches the underlying data's arbitrage profile cell for
cell: on BTC neither the data nor the ConvVAE reconstructions
produce any calendar or butterfly violation at any mask rate, and
on ETH the ConvVAE inherits the same $0.2\%$ calendar-violation rate
as the gridded data (traceable to a small number of high-spread,
low-liquidity snapshots) while remaining butterfly-free. The hybrid
inherits the same property because the smile re-fit operates only
on rows with $\ge 3$ observed cells, where the parametric fit is
well-determined and itself near-compliant. Second, the calendar
projection step is operationally free: it changes hidden-cell RMSE
by at most $0.001$ vol points on the learned predictors, and in
fact \emph{improves} smile-refit RMSE by up to $0.2$ vol points on
its own outputs. Third, the smile re-fit alone is the only
predictor that materially violates either condition at high mask
rates: $38.9\%$ ($33.1\%$) of its BTC (ETH) reconstructions at
$r = 0.5$ admit a butterfly arbitrage at the listed strikes. The
routing rule of Section~\ref{sec:hybrid} suppresses this failure
mode in the hybrid by deferring to the ConvVAE on rank-deficient
rows.

Static no-arbitrage at the seven listed strikes is therefore a
demonstrated property of the deployed predictor at no measurable
accuracy cost. Arbitrage \emph{between} the listed strikes requires
a smooth interpolant \citep{gatheral2014svi, fengler2007arbitragefree}
and is the one no-arbitrage condition this paper does not address.

% =============================================================================
\section{Structured Holes}
\label{sec:structured}

Random per-cell masking is the standard evaluation in the machine
learning literature, but the distribution of missing cells in
production systems is rarely independent across the surface. A feed
disruption typically removes an entire tenor row, and the withdrawal
of a market maker from one wing removes a delta column. We therefore
evaluate on five fixed-pattern masks corresponding to such scenarios
and compare each against a random-mask control of identical cell
count, so that the effect of mask \emph{structure} is isolated from
the effect of mask \emph{rate}.

\begin{table}[!htbp]
\centering
\caption{Structured-hole evaluation (vol points), joint-trained ConvVAE
($z{=}16, h{=}64$). ``struct.''\ is RMSE under the structured mask;
``rnd.''\ is RMSE under a random mask hiding the same number of
cells. ``Hid.''\ is cells hidden per snapshot.}
\label{tab:structured}
\small
\setlength{\tabcolsep}{4pt}
\renewcommand{\arraystretch}{1.05}
\begin{tabular}{lccccccc}
\toprule
 & & & \multicolumn{2}{c}{Smile re-fit} & \multicolumn{2}{c}{ConvVAE} & Hyb. \\
\cmidrule(lr){4-5} \cmidrule(lr){6-7}
Scheme & Hid. & Rate & struct. & rnd. & struct. & rnd. & struct. \\
\midrule
\texttt{row\_random}  &  7 & 0.17 & 9.59 & 0.14 & 1.88 & 1.03 & 1.88 \\
\texttt{col\_random}  &  6 & 0.14 & 0.00 & 0.00 & 0.94 & 1.00 & 0.00 \\
\texttt{wing\_put}    & 12 & 0.29 & 0.00 & 2.22 & 1.50 & 1.05 & 0.00 \\
\texttt{wing\_call}   & 12 & 0.29 & 0.00 & 2.22 & 1.20 & 1.05 & 0.00 \\
\texttt{long\_tenor}  &  7 & 0.17 & 13.12 & 0.14 & 1.54 & 1.03 & 1.54 \\
\bottomrule
\end{tabular}
\end{table}

\begin{figure*}[!htbp]
\centering
\includegraphics[width=0.92\linewidth]{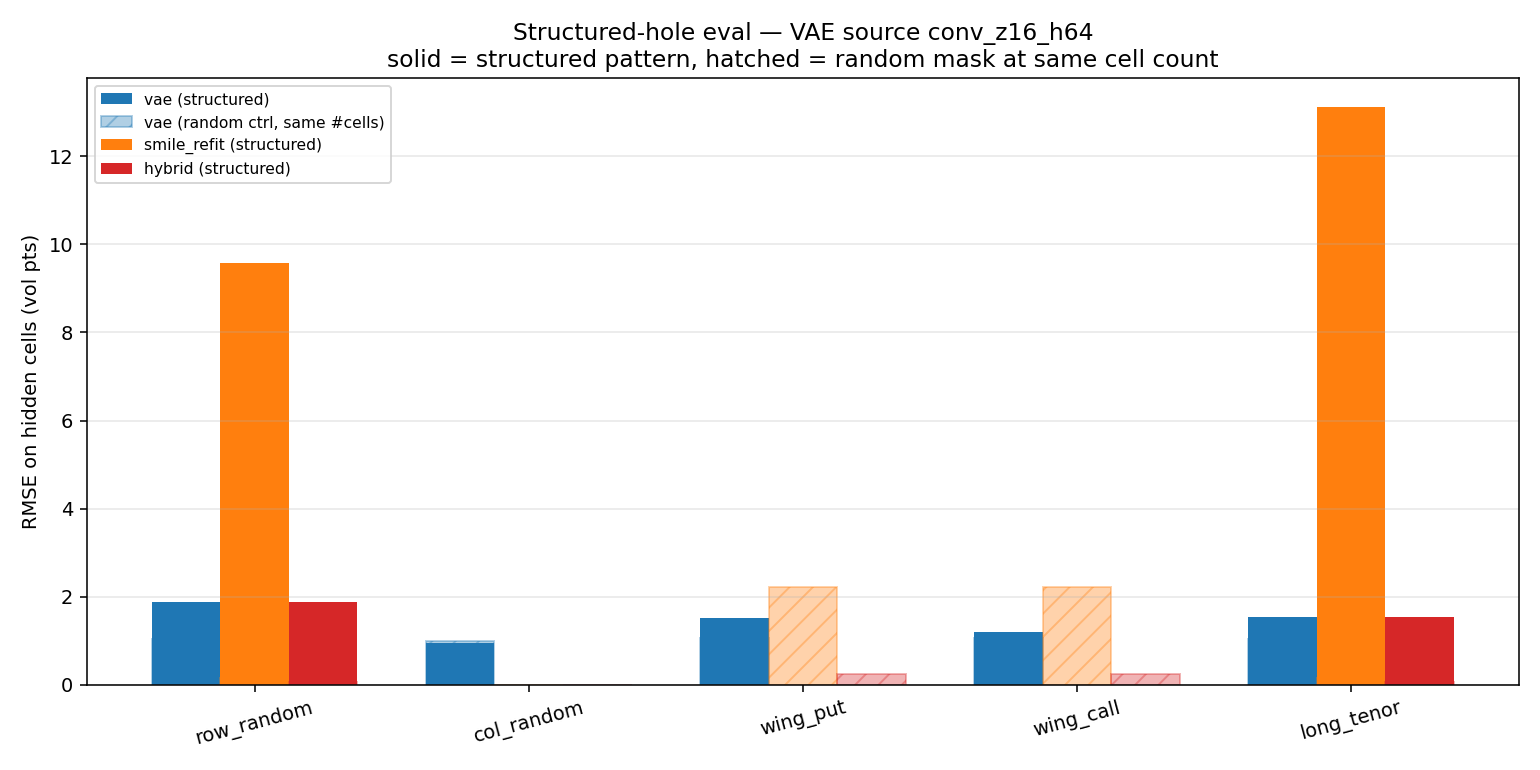}
\caption{Structured-hole evaluation. Solid bars: RMSE under the
structured mask. Hatched bars: RMSE under a random mask with the
same number of hidden cells. The smile re-fit collapses on
row-shaped holes (\texttt{row\_random}, \texttt{long\_tenor}) where
one tenor has zero observed cells; the ConvVAE remains usable.
Where a tenor row retains $\ge 3$ observed cells
(\texttt{col\_random}, wings), the smile re-fit is essentially
perfect because the gridded targets are outputs of the same
parametric family, and the hybrid correctly routes to it.}
\label{fig:structured}
\end{figure*}

Table~\ref{tab:structured} and Figure~\ref{fig:structured} partition
the scenarios into two regimes. In the column-hole and wing-hole scenarios every tenor row
retains at least five of seven cells; the smile re-fit is
rank-sufficient at each tenor and reproduces the gridded targets to
within numerical tolerance, while the ConvVAE incurs $0.9$--$1.5$
vol points; the hybrid defers to the re-fit. In the row-hole
scenarios (\texttt{row\_random}, \texttt{long\_tenor}) the re-fit is
rank-deficient at the affected tenor, forced to cross-tenor
extrapolation of $(a, b, c)$, and incurs $9.6$--$13.1$ vol
points, an order of magnitude above the random-mask control. The
ConvVAE attains $1.54$--$1.88$ vol points in the same regime, and
the hybrid defers to it. Comparison of the ConvVAE under structured versus random masks at
matched cell counts shows a residual distribution-shift cost on
\texttt{row\_random}: $1.88$ vs.\ $1.03$ vol points, a $1.83\times$
penalty. The residual is bounded and substantially smaller than the
order-of-magnitude failure of the parametric baseline in the same
scheme.

% =============================================================================
\section{Cross-Market Generalisation}
\label{sec:cross_market}

Whether a single model can serve more than one cryptocurrency
market is tested by two complementary experiments: zero-shot
out-of-distribution evaluation of a BTC-only ConvVAE on the ETH
test set, and joint training on both symbols with per-market
evaluation.

\subsection{Zero-shot out-of-distribution transfer}
\label{sec:transfer}

We take the BTC-only ConvVAE and evaluate it directly on the ETH
test set, normalising the inputs with ETH's own training statistics,
which is the realistic deployment configuration. As a complementary
diagnostic we also evaluate the same model with normalisation drawn
from the source's (BTC's) training statistics, which separates the
contribution of shape transfer from any coincidence in marginal IV
levels. Results are reported in Table~\ref{tab:cross_asset}.

\begin{table*}[!htbp]
\centering
\caption{Zero-shot transfer: BTC-only ConvVAE evaluated on the ETH
test set, against the in-distribution BTC reference and the
ETH-specific smile re-fit baseline (vol points).}
\label{tab:cross_asset}
\vspace{6pt}
\small
\setlength{\tabcolsep}{10pt}
\begin{tabular}{lccccc}
\toprule
Mask rate & $r = 0.10$ & $r = 0.20$ & $r = 0.30$ & $r = 0.40$ & $r = 0.50$ \\
\midrule
BTC test (in-distribution reference)   & 1.28 & 1.39 & 1.51 & 1.47 & 1.67 \\
ETH test, target normalisation         & 1.63 & 1.61 & 1.58 & 1.66 & 1.82 \\
ETH test, source normalisation         & 1.48 & 1.45 & 1.44 & 1.49 & 1.66 \\
\midrule
ETH smile re-fit baseline              & 0.00 & 0.83 & 1.52 & 3.71 & 5.84 \\
\bottomrule
\end{tabular}
\end{table*}

Under target-symbol normalisation the cross-asset RMSE on ETH is
within $5$--$27\%$ of the in-distribution BTC reference across the
five mask rates evaluated. At $r = 0.5$ the BTC-trained ConvVAE on
ETH attains $1.82$ vol points, $3.2\times$ below the ETH-specific
parametric baseline ($5.84$). Source-symbol normalisation is in
fact marginally more accurate than target-symbol normalisation at
the high-mask end ($1.66$ vs.\ $1.82$ vol points at $r = 0.5$),
consistent with the close alignment of the per-cell IV
distributions of BTC and ETH over the window: the $60$-day ATM IV
means of the two symbols differ by only $1.1$ vol points. The
representation learned by the ConvVAE is therefore not BTC-specific,
and the cryptocurrency vol-surface manifold within the observation
window is substantially shared between the two largest markets.

\subsection{Joint training}
\label{sec:joint}

\begin{figure*}[!htbp]
\centering
\includegraphics[width=0.92\linewidth]{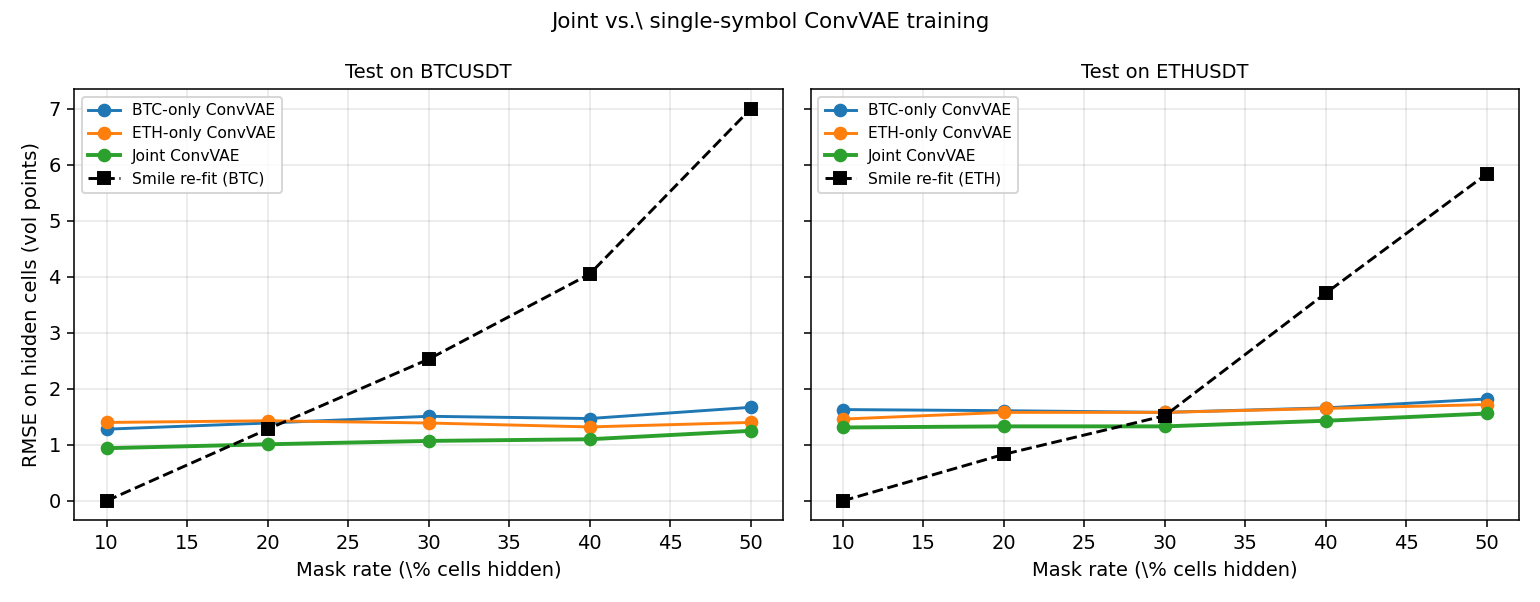}
\caption{Joint vs.\ single-symbol ConvVAE training. The joint model
is the lowest-error learned predictor on each market at every mask
rate, including the in-distribution market on which each
single-symbol model was specifically trained.}
\label{fig:joint}
\end{figure*}

We additionally train a ConvVAE on the concatenation of per-symbol
$z$-normalised BTC and ETH training surfaces, yielding a
$4{,}223$-surface training set against the $1{,}974$ and $2{,}249$
surfaces used for the single-symbol counterparts. The architecture
and optimiser are unchanged. Each market's test set is evaluated
under its own normalisation; results appear in
Table~\ref{tab:joint} and Figure~\ref{fig:joint}.

\begin{table}[!htbp]
\centering
\caption{Joint vs.\ single-symbol ConvVAE training (vol points). Bold
marks the lowest value in each column. The smile baseline wins at
low mask rates (where it is the parametric oracle on the gridded
targets) and the joint ConvVAE wins at higher mask rates. Among
learned models, the joint ConvVAE is uniformly best on both
markets at every mask rate.}
\label{tab:joint}
\small
\begin{tabular}{lccccc}
\toprule
 & $0.10$ & $0.20$ & $0.30$ & $0.40$ & $0.50$ \\
\midrule
\multicolumn{6}{l}{\emph{BTC test set}} \\
BTC-only ConvVAE       & 1.28 & 1.39 & 1.51 & 1.47 & 1.67 \\
ETH-only ConvVAE       & 1.40 & 1.43 & 1.39 & 1.32 & 1.40 \\
Joint ConvVAE          & 0.94 & 1.01 & \textbf{1.07} & \textbf{1.13} & \textbf{1.25} \\
BTC smile baseline     & \textbf{0.00} & \textbf{1.28} & 2.53 & 4.05 & 7.00 \\
\midrule
\multicolumn{6}{l}{\emph{ETH test set}} \\
BTC-only ConvVAE       & 1.63 & 1.61 & 1.58 & 1.66 & 1.82 \\
ETH-only ConvVAE       & 1.46 & 1.58 & 1.58 & 1.65 & 1.72 \\
Joint ConvVAE          & 1.31 & 1.33 & 1.33 & \textbf{1.43} & \textbf{1.56} \\
ETH smile baseline     & \textbf{0.00} & \textbf{0.83} & \textbf{1.52} & 3.71 & 5.84 \\
\bottomrule
\end{tabular}
\end{table}

The joint ConvVAE is the lowest-error learned predictor on both
test sets at every mask rate. Relative to the better-performing
single-symbol counterpart it reduces error by $9$--$27\%$ across
the ten test-set/mask-rate combinations. Against the matching
in-distribution single-symbol model the reduction is larger on
BTC ($23$--$29\%$, joint vs.\ BTC-only) than on ETH ($9$--$16\%$,
joint vs.\ ETH-only), reflecting that the BTC training set is
both smaller and structurally less clean than the ETH one ($1{,}974$
vs.\ $2{,}249$ fully-filled snapshots, $80.9\%$ vs.\ $92.1\%$
fully-filled fraction; Section~\ref{sec:data}). The ETH-only
ConvVAE in turn exceeds the BTC-only ConvVAE on the BTC test set at
high mask rates ($1.40$ vs.\ $1.67$ at $r = 0.5$); at our sample
size training-set quality dominates the nominal in-distribution
advantage. Both the increase in effective training-set size and
the additional shape diversity introduced by the second market
contribute to the joint gain.

\subsection{Synthesis across the evaluation grid}
\label{sec:synthesis}

\begin{figure*}[t]
\centering
\includegraphics[width=\linewidth]{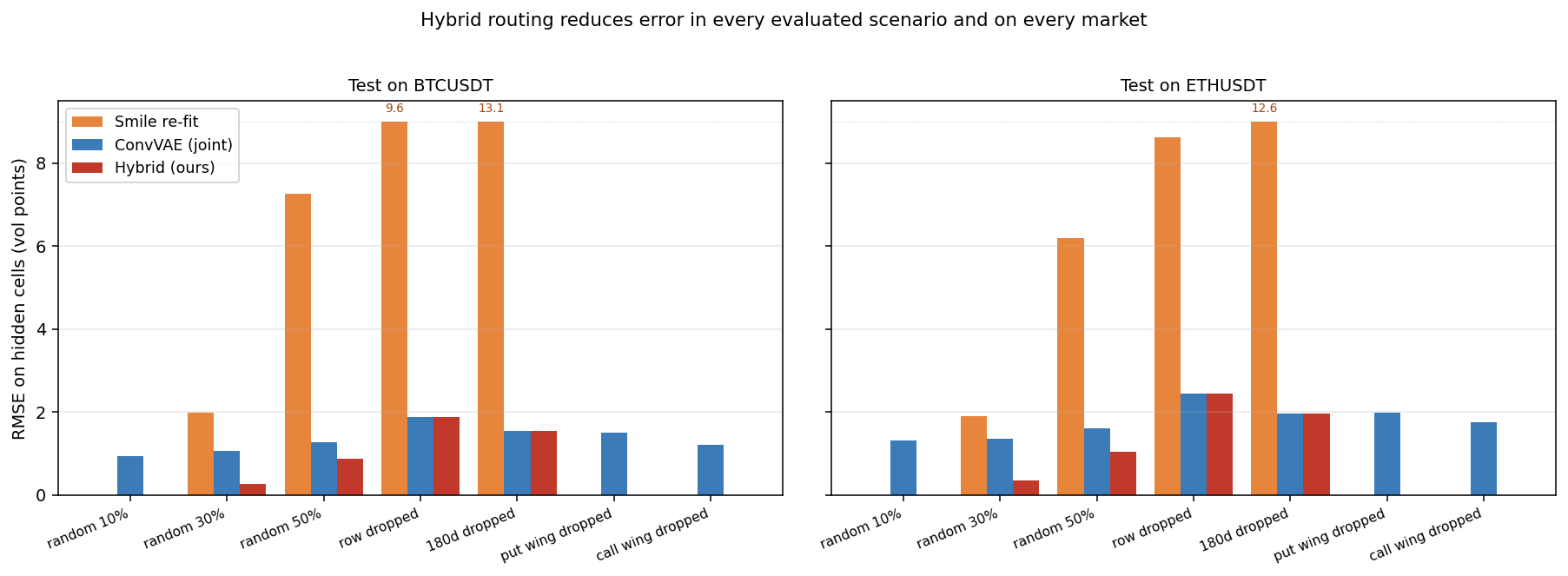}
\caption{Hidden-cell RMSE of the smile re-fit (orange), the joint
ConvVAE (blue), and the hybrid (red) across seven evaluation
scenarios on the BTC and ETH test sets. Bars are clipped at $9$ vol
points for legibility; the true value is annotated above any clipped
bar. In every scenario on every market, the hybrid attains the
lowest error. The smile re-fit is competitive only when each tenor
row retains $\ge 3$ observed cells (wing- and column-hole scenarios);
the ConvVAE is the only viable predictor when a tenor row is fully
unobserved.}
\label{fig:summary}
\end{figure*}

Figure~\ref{fig:summary} consolidates the combined effect of the
routing rule, the joint model, and cross-market training across the
seven evaluation scenarios examined in this paper: three random-mask
rates ($10\%$, $30\%$, $50\%$) and four structured holes (an entire
tenor row dropped, the longest tenor dropped, the put wing dropped,
and the call wing dropped), evaluated separately on the BTC and ETH test sets.
The hybrid attains the lowest error in every scenario on both
markets. In the wing-hole scenarios the smile re-fit is
rank-sufficient and the hybrid inherits its near-zero error; in the
random-mask and row-hole scenarios the joint ConvVAE provides the
fallback, keeping the hybrid in low single-digit vol points
everywhere even as the smile re-fit alone exceeds twelve. The
hybrid Pareto-dominates each component across the entire evaluation
grid.

% =============================================================================
\section{Anomaly Detection Case Study}
\label{sec:anomaly}

The joint ConvVAE trained for masked completion in
Sections~\ref{sec:completion}--\ref{sec:cross_market} also yields an
unsupervised per-snapshot anomaly score. With the mask set to
all-observed, the reconstruction error $\| x - \hat x \|$ quantifies
the distance of a surface from the manifold that the model has been
trained to encode. We compute this score for every BTC snapshot in
the train, validation, and test windows of the $147$-day record and
examine the resulting time series and latent geometry.

Across the $2{,}821$ scored BTC snapshots the mean reconstruction
RMSE is $0.76$ vol points; the $99$th percentile is $2.17$; the
maximum is $4.30$. The top five anomalies fall in late September
and October $2023$, with timestamps \texttt{2023-10-02 03:00}
($4.30$), \texttt{2023-10-23 10:00} ($3.24$),
\texttt{2023-09-29 18:00} ($2.62$), \texttt{2023-10-20 02:00}
($2.59$), and \texttt{2023-09-29 19:00} ($2.57$ vol points).

\begin{figure*}[t]
\centering
\includegraphics[width=0.95\linewidth]{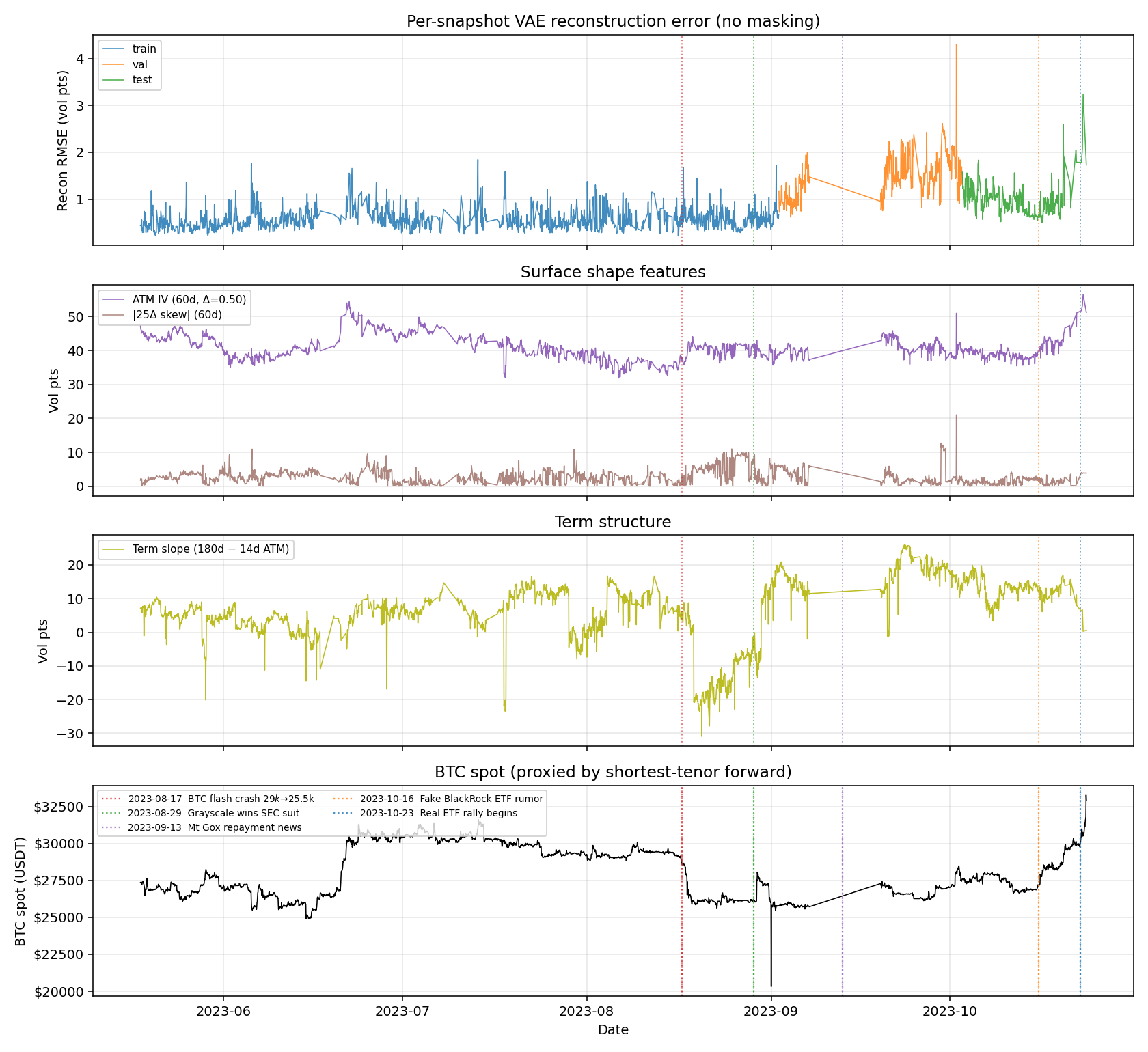}
\caption{Anomaly forensics. Top: per-snapshot reconstruction RMSE
under no masking, with train/val/test windows distinguished. Second
and third: diagnostic surface features (ATM, skew, term slope).
Bottom: BTC spot proxied by shortest-tenor parity-implied forward.
Dotted vertical lines mark known events.}
\label{fig:anomaly_timeline}
\end{figure*}

Figure~\ref{fig:anomaly_timeline} overlays the reconstruction-error
time series on three diagnostic surface features: the at-the-money
($\delta = 0.50$) implied volatility at the $60$-day tenor, the
$25$-delta skew at the same tenor, and the term-structure slope
between the $14$- and $180$-day at-the-money points, together with
the BTC spot price proxied by the shortest-tenor parity-implied
forward. Several features of the period are evident. The
at-the-money volatility declines from approximately $55\%$ at the
start of the window to approximately $25\%$ by early September, a
level shift of $\sim$$30$ volatility points; the term structure
inverts and becomes irregular through September; and the spot price
rises from $\$26{,}000$ in mid-October to $\$33{,}000$ by month end.
The early-October peak (October $2$) precedes the late-October
ETF-anticipation rally by roughly two weeks, and the rally itself
appears as the cluster of high-error snapshots on October $20$
and $23$. The August $17$ flash crash (an intraday move from
$\$29{,}000$ to $\$25{,}500$) is visible as a localised spike in
reconstruction error rather than a top-ranked anomaly. The
September $29$ cluster does not coincide with a major
crypto-specific event known to the authors; the model identifies
it from the surface data alone.

\begin{figure*}[!htbp]
\centering
\includegraphics[width=0.82\linewidth]{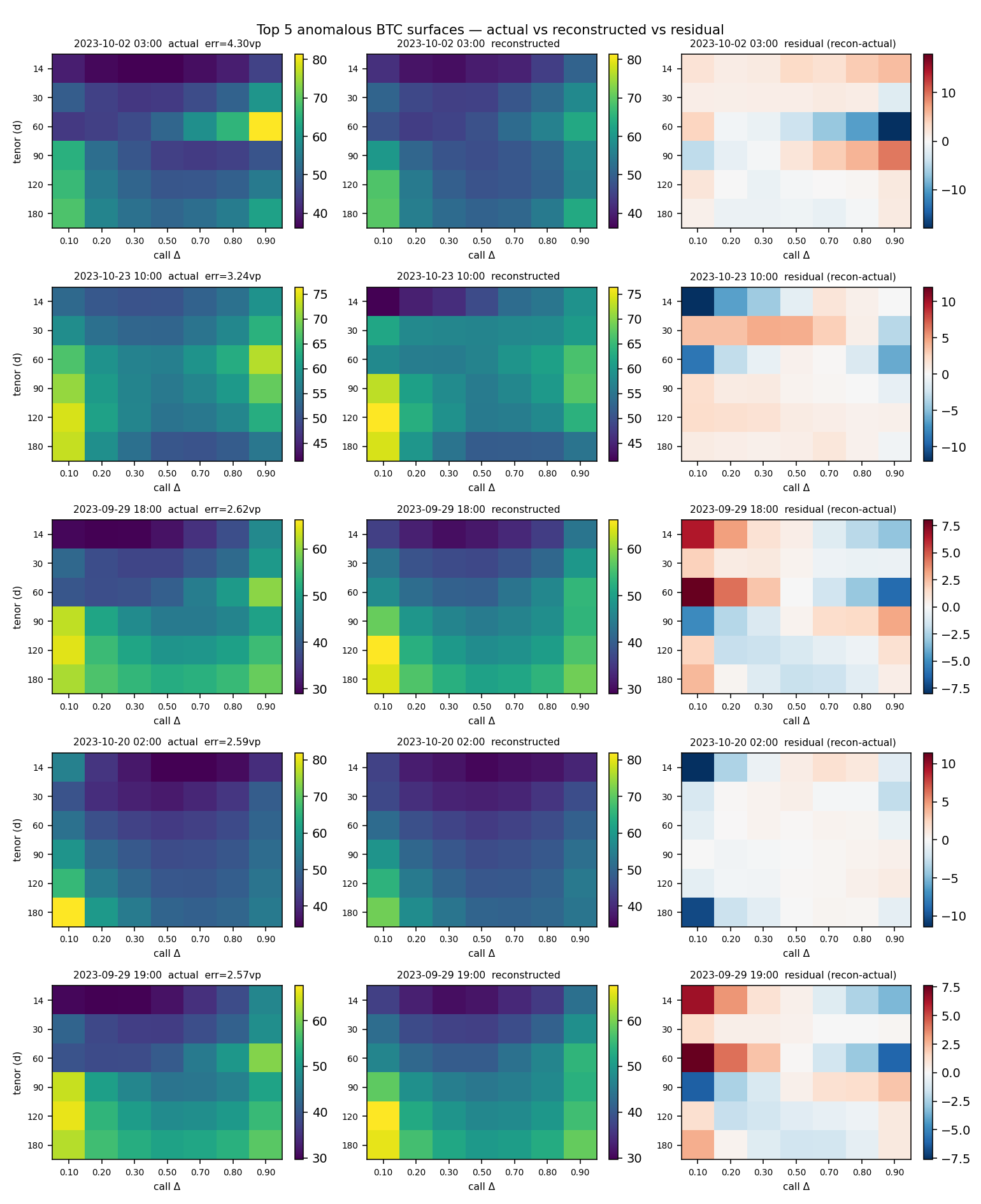}
\caption{Top-$5$ anomalous BTC surfaces by ConvVAE reconstruction
error. Each row shows the actual surface (left), the ConvVAE
reconstruction (centre), and the signed residual (right; red
$=$ over-prediction, blue $=$ under-prediction). Residuals are
spatially coherent rather than random, indicating genuine
off-manifold structure rather than fitted noise.}
\label{fig:anomaly_surfaces}
\end{figure*}

Examination of the corresponding surfaces
(Figure~\ref{fig:anomaly_surfaces}) shows that the high-error
snapshots exhibit systematic, spatially-coherent residuals rather
than random fluctuations: the residual heatmaps display
sign-coherent blocks in the deep out-of-the-money call wing and in
the short-tenor at-the-money region, indicating that the model is
failing to reproduce genuine local shape features rather than
overfitting to noise. This is consistent with these snapshots
residing at the periphery of the learned manifold rather than off
it.

\begin{figure}[!htbp]
\centering
\includegraphics[width=0.85\linewidth]{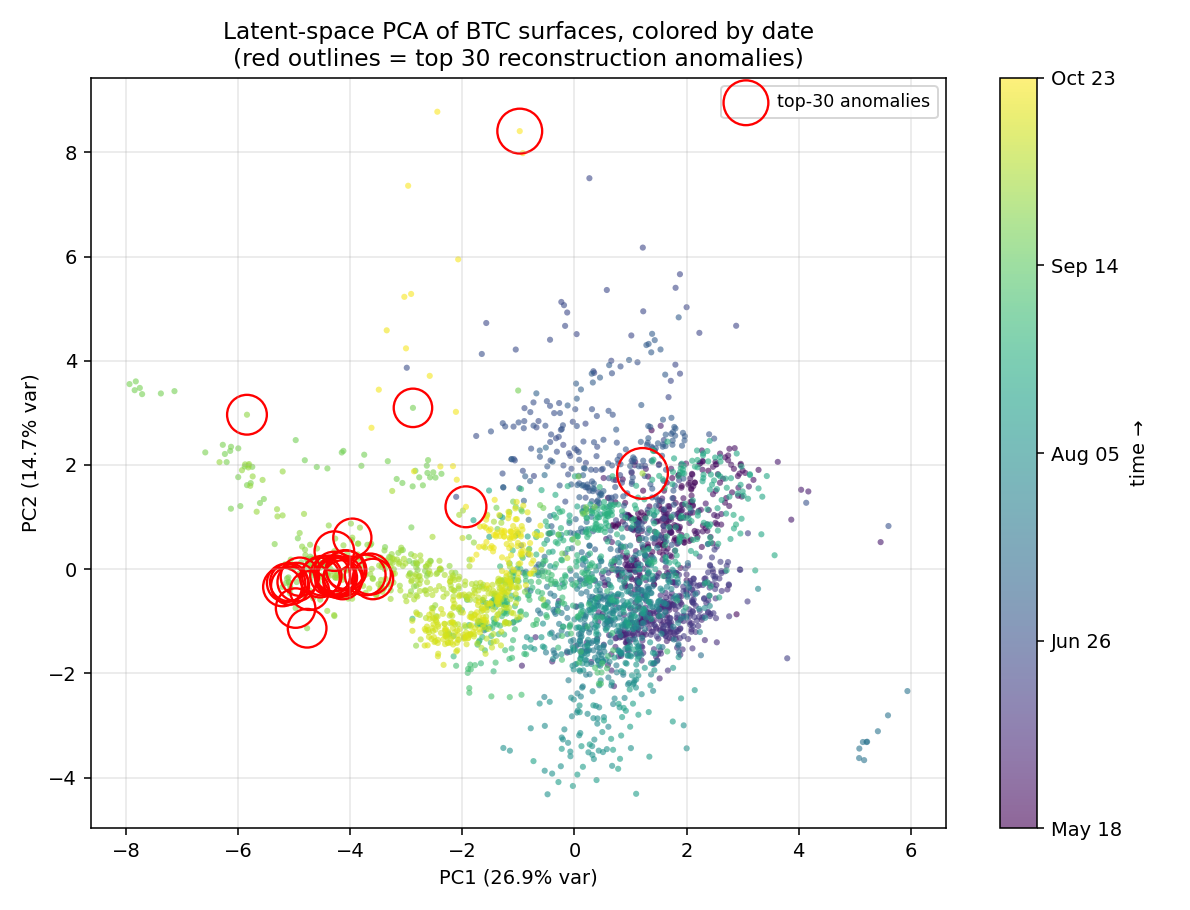}
\caption{Two-dimensional PCA projection of the $z$-dim latent means.
Colour: time. Red rings: top-$30$ reconstruction anomalies.
Anomalies sit at the manifold edges, not within the dense interior.}
\label{fig:anomaly_latent}
\end{figure}

Projection of the encoded latent means $\boldsymbol\mu_i$ onto their
first two principal components (Figure~\ref{fig:anomaly_latent})
reveals interpretable manifold structure. The snapshots trace a
continuous temporal trajectory in the $(\text{PC1}, \text{PC2})$
plane between the start and end of the $147$-day window, with the
high- and low-volatility regimes occupying separable regions. The
top-$30$ anomalies (red rings) concentrate at the periphery of the
dense scatter rather than within its interior, the expected
qualitative signature of a generative-model anomaly score on a
well-trained representation.

The anomaly result requires no labels, no re-training, and no
threshold beyond a ranking of per-snapshot errors. It is an
analytical by-product of a trained surface model rather than a
complete anomaly-detection study; a complete study would compare
against labelled events, drift baselines, and sliding-window
statistical detectors. The claim is the narrower one: the
representation learned for masked completion is also informative
for flagging surfaces that warrant further investigation.

% =============================================================================
\section{Deployment Considerations}
\label{sec:discussion}

The empirical results of
Sections~\ref{sec:arch}--\ref{sec:anomaly} support a single
deployable configuration. A predictor intended to serve a
multi-market cryptocurrency options book should consist of a $2$D
convolutional VAE (trained jointly on the per-symbol $z$-normalised
concatenation of all available training surfaces) queried through
the routing rule of Eq.~\eqref{eq:hybrid}: the smile re-fit when
each tenor row retains at least three observed cells, and the
ConvVAE otherwise. The same trained model provides, at no
additional inference cost, a per-snapshot reconstruction-error
statistic that may be used to flag surfaces lying away from the
learned manifold, and produces calendar- and butterfly-arbitrage-free
outputs at the listed strikes by construction
(Section~\ref{sec:arb}).

The computational requirements of the configuration are modest.
Both predictors execute in time that is independent of chain depth:
the ConvVAE consists of $318$k parameters and reduces at inference
to a fixed sequence of small convolutional passes over the
$6 \times 7$ grid, while
the smile re-fit reduces to a single ordinary least-squares solve
per tenor on a system of at most seven points. Neither requires
accelerator hardware at inference, and both can be hosted within a
common surface-construction service.

The marginal contribution of the ConvVAE is concentrated in the
regime in which the parametric smile is rank-deficient, namely tenor rows
with fewer than three observed cells, as characterised in
Section~\ref{sec:structured}. In the complementary regime, where
the parametric smile is well-determined, the gridded targets impose
an upper bound on the accuracy attainable by any learned model and
the routing rule defers to the parametric predictor by
construction. The cross-market results of
Section~\ref{sec:cross_market} provide a complementary justification
for the modelling cost: jointly trained models outperform
single-symbol alternatives on every test market examined, including
the symbol on which each single-symbol model was specifically
trained. Both an increase in effective training-set size and the
additional shape diversity introduced by a second market plausibly
contribute to the gain.

% =============================================================================
\section{Limitations and Future Work}
\label{sec:limitations}

\paragraph{Window length and regime coverage.}
Our empirical record spans the $147$-day Binance Options EOH archive
(May--October $2023$). The window contains one major intra-period
dislocation (the August $17$ flash crash) and the late-October
ETF-anticipation rally, but does not span structurally distinct
regimes such as the FTX collapse, the $2024$ spot-ETF launch, or
post-halving environments. The cross-asset transfer results of
Section~\ref{sec:cross_market} should accordingly be interpreted as
evidence of a shared BTC--ETH manifold over this window, not as a
guarantee of generalisation across the full crypto-volatility regime
space.

\paragraph{Gridded targets and the parametric oracle.}
The training and evaluation targets are themselves the output of a
parametric gridding procedure (Section~\ref{sec:grid}), which makes
the smile re-fit baseline the inverse of the data-generating map
and unbeatable whenever each tenor row is adequately populated.
Training the ConvVAE directly on raw cleaned chains, bypassing the
parametric gridding, would relax this asymmetry but would require a
permutation-equivariant or set-structured encoder and is left as an
extension.

\paragraph{Residual structured-vs-random penalty.}
The ConvVAE retains a $1.83\times$ structured-vs-random penalty on
\texttt{row\_random}. Closing it further requires an
encoder--decoder that explicitly conditions each tenor on its
observed neighbours (hierarchical latents indexed by tenor, or
arbitrage-constrained decoder factorisations) rather than relying
on convolutional receptive-field growth alone.

\paragraph{Continuous-smile arbitrage.}
Section~\ref{sec:arb} establishes that the deployed predictor is
calendar- and butterfly-arbitrage-free at the seven listed strikes
per tenor: calendar arbitrage via a free isotonic post-projection, butterfly
empirically without enforcement. Arbitrage \emph{between} the
listed strikes (for example, at a strike interpolated for a Greek
on an off-grid expiry) requires a smooth interpolant: an
arbitrage-free SVI \citep{gatheral2014svi} or a Fengler-style
$C^2$ spline \citep{fengler2007arbitragefree}. We do not address
this finer condition; pairing the ConvVAE outputs with a
constrained interpolant is the natural next step.

\paragraph{Cross-asset-class transfer.}
We address only cryptocurrency markets. Whether the same
architecture transfers to equity-index options, which exhibit
pronounced left-skew and qualitatively different term-structure
dynamics than crypto, requires a separate treatment of the
corresponding data pipeline and arbitrage constraints and is
beyond the scope of this paper.

% =============================================================================
\section{Conclusion}
\label{sec:conclusion}

A $2$D-convolutional masked-input VAE for the cryptocurrency
volatility surface, combined with a quadratic smile re-fit through
a deterministic per-tenor routing rule, attains the lowest
hidden-cell RMSE at every random and structured masking scenario
examined and on both BTC and ETH test sets
(Figure~\ref{fig:summary}). At $50\%$ random masking the hybrid
attains $0.83$ vol points against $7.00$ for the smile re-fit
alone, an eightfold reduction over standard parametric practice
obtained at no additional inference cost.

The hybrid additionally removes a categorical failure mode of the
parametric baseline. When a tenor row is fully unobserved (a
configuration routinely produced in production by feed failures or
maturity delistings), the smile re-fit is rank-deficient at the
affected tenor and incurs $9.6$--$13.1$ vol points of error
(``row dropped'' and ``180d dropped'' panels of
Figure~\ref{fig:summary}); the ConvVAE retains $1.5$--$1.9$ vol
points and the routing rule defers to it automatically.

The cryptocurrency vol-surface manifold is substantially shared
across the two largest markets over the observation window: a
ConvVAE trained on BTC alone attains within $5$--$27\%$ of its
in-distribution accuracy on ETH, and joint training on BTC and ETH
yields a further $9$--$27\%$ reduction on every market
examined, including the symbol on which each single-symbol model
was specifically trained. A single jointly-trained ConvVAE is
therefore the appropriate choice for a multi-currency portfolio.

The deployed predictor is calendar- and butterfly-arbitrage-free at
the seven listed strikes per tenor on both markets: calendar via a
free $L_2$ post-projection that moves hidden-cell RMSE by at most
$0.001$ vol points, and butterfly empirically without enforcement
(Section~\ref{sec:arb}). The parametric smile re-fit, by contrast,
admits a butterfly arbitrage at the listed strikes on $38.9\%$
($33.1\%$) of BTC (ETH) reconstructions at $50\%$ masking; the
routing rule suppresses this failure mode in the hybrid.

The same trained model yields, at no additional cost and without
supervision, a per-snapshot reconstruction-error statistic that
flags the late-October ETF-anticipation rally and the August $17$,
$2023$ flash crash as elevated-error periods, and a latent
representation that traces an interpretable temporal trajectory. All
training and evaluation infrastructure is released to support
reproducible follow-on work.

\paragraph{Code and Data Availability.}
The code for the data pipeline, model training, the ablation study,
and all figures and tables in this manuscript is available at
\url{https://github.com/jasper-research/beyond-the-smile-paper} under
the MIT License. An archived snapshot of the code, together with the
processed $6 \times 7$ gridded volatility surfaces and the per-run
configurations, checkpoints, and metric files underlying the reported
results, is deposited on Zenodo
(DOI:~\href{https://doi.org/10.5281/zenodo.20693546}{\texttt{10.5281/zenodo.20693546}}).
The sole data source, the Binance Options end-of-hour archive, is
publicly available. The complete ablation grid of
Sections~\ref{sec:completion}--\ref{sec:joint} trains and evaluates in
under five minutes of GPU time on commodity hardware, permitting full
re-verification of the reported results from the raw archive.

% =============================================================================
\bibliographystyle{plainnat}
\bibliography{references}

\end{document}